\documentclass[10pt,twocolumn,letterpaper]{article}

\usepackage{cvpr}
\usepackage{times}
\usepackage{epsfig}
\usepackage{graphicx}
\usepackage{amsmath}
\usepackage{amssymb}
\usepackage{bm}
\usepackage{caption}
\usepackage{booktabs}
\usepackage[page,header]{appendix}
\usepackage{multibib}
\newcommand{\aksoy}{Aksoy {\it et al.} }
\newcommand{\koyama}{Koyama {\it et al.} }
\newcommand{\jtan}{Tan {\it et al.} } 

\usepackage[symbol]{footmisc}

\usepackage[breaklinks=true,bookmarks=false]{hyperref}

\cvprfinalcopy 

\def\cvprPaperID{7640} 

\ifcvprfinal\fi

\begin{document}

\title{ Fast Soft Color Segmentation}

\author{Naofumi Akimoto$^1$ \footnote[1] \quad \quad \quad Huachun Zhu$^2$ 
	\quad \quad \quad  Yanghua Jin$^2$ \quad \quad \quad  Yoshimitsu Aoki$^1$ \\
	$^1$ Keio University \quad $^2$ Preferred Networks\\
	{\tt\small nakimoto@aoki-medialab.jp \quad \{zhu, jinyh\}@preferred.jp \quad aoki@elec.keio.ac.jp}
}

\twocolumn[{%
\renewcommand\twocolumn[1][]{#1}%
\maketitle
\begin{center}
\centering
\includegraphics[width=\textwidth]{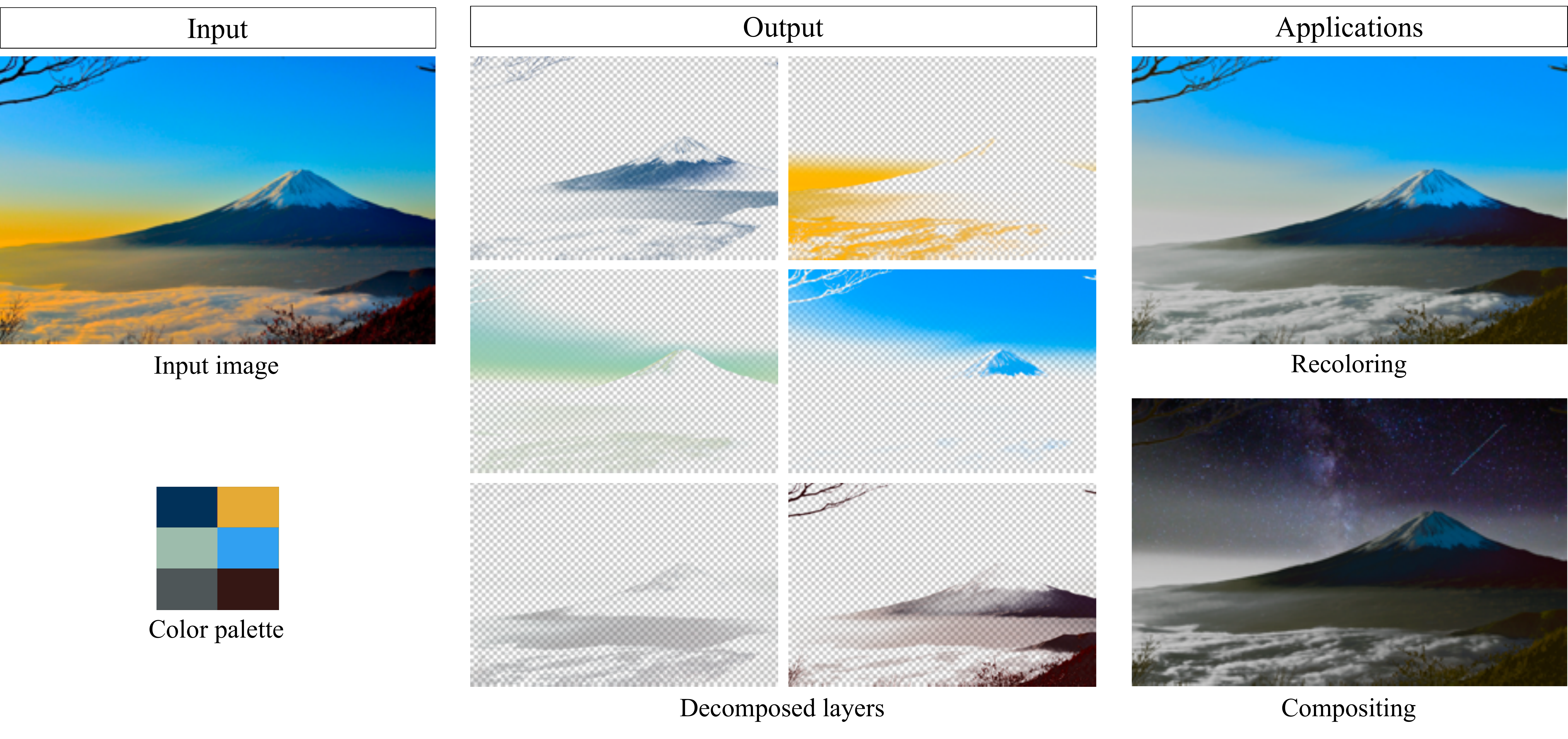}
\captionof{figure}{
    We propose a neural network based framework that, by using a single RGB image and a specified color palette, decomposes the image into multiple RGBA layers, each of which contains homogeneous colors. The decomposed layers can be created instantaneously and are useful for image and video editing, such as recoloring or compositing.
    }
\label{fig:teaser}
\end{center}%
}]

\footnotetext[1]{A part of this work was done while the first author worked at Preferred Networks as an intern.}


\begin{abstract}
We address the problem of soft color segmentation, defined as decomposing a given image into several RGBA layers, each containing only homogeneous color regions.
The resulting layers from decomposition pave the way for applications that benefit from layer-based editing, such as recoloring and compositing of images and videos. 
The current state-of-the-art approach for this problem is hindered by slow processing time due to its iterative nature, and consequently does not scale to certain real-world scenarios. 
To address this issue, we propose a neural network based method for this task that decomposes a given image into multiple layers in a single forward pass. 
Furthermore, our method separately decomposes the color layers and the alpha channel layers. By leveraging a novel training objective, our method achieves proper assignment of colors amongst layers.
As a consequence, our method achieve promising quality without existing issue of inference speed for iterative approaches.
Our thorough experimental analysis shows that our method produces qualitative and quantitative results comparable to previous methods while achieving a 300,000x speed improvement.
Finally, we utilize our proposed method on several applications, and demonstrate its speed advantage, especially in video editing.
\end{abstract}


\begin{figure*}[t]
\begin{center}
\includegraphics[width=\linewidth]{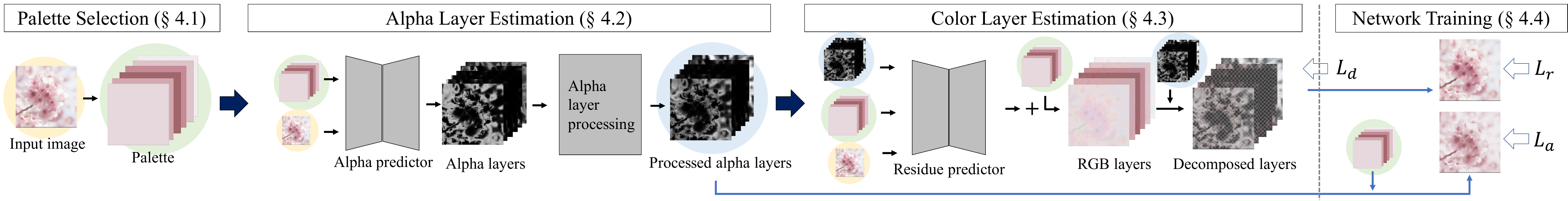}

   \caption{Our framework consists of three stages. First, palette colors are selected manually or automatically based on an input image. Second, an alpha predictor (U-Net) predicts alpha layers using the input image and the palette. Then, the alpha layers are normalized to satisfy an alpha-add condition (Eq. (\ref{eq:alpha-unity})), or undergo an alpha layer processing, such as smooth filtering or mask operation. Finally, with the input image, the palette, and the processed alpha layers, a residue predictor (U-Net) estimates the difference between the palette color and the ground-truth image at each layer in order to restore the input image from the decomposed layers. These two networks are trained jointly.}
\label{fig:system}
\end{center}
\end{figure*}

\section{Introduction}
Image segmentation is the task of decomposing an image into meaningful regions.
A typical approach to this problem assigns a single class to each pixel in an image. 
However, such hard segmentation is far from ideal when the distinction of meaningful regions is ambiguous, such as objects with motion blur, transparency or illumination.
A flexible alternative is soft segmentation, which allow a single pixel to be assigned to multiple classes.
This paper addresses a subtask of soft segmentation: soft color segmentation.
the goal of this task is to decompose an RGB image into several color-considering RGBA layers, each of which is made up of homogeneous colors.
The decomposed soft color layers allow a user to target specific color regions in editing tasks such as recoloring and compositing.
In an approach to solve the problem, existing methods adopt either an optimization-based \cite{aksoy16,aksoy17,koyama} or a geometric \cite{tan18, tan16} approach. 
However, it takes a considerable amount of time for existing methods to decompose a high-resolution image or a series of images (e.g. frames in a video).

\begin{figure}[t]
\begin{center}
\includegraphics[width=0.95\linewidth]{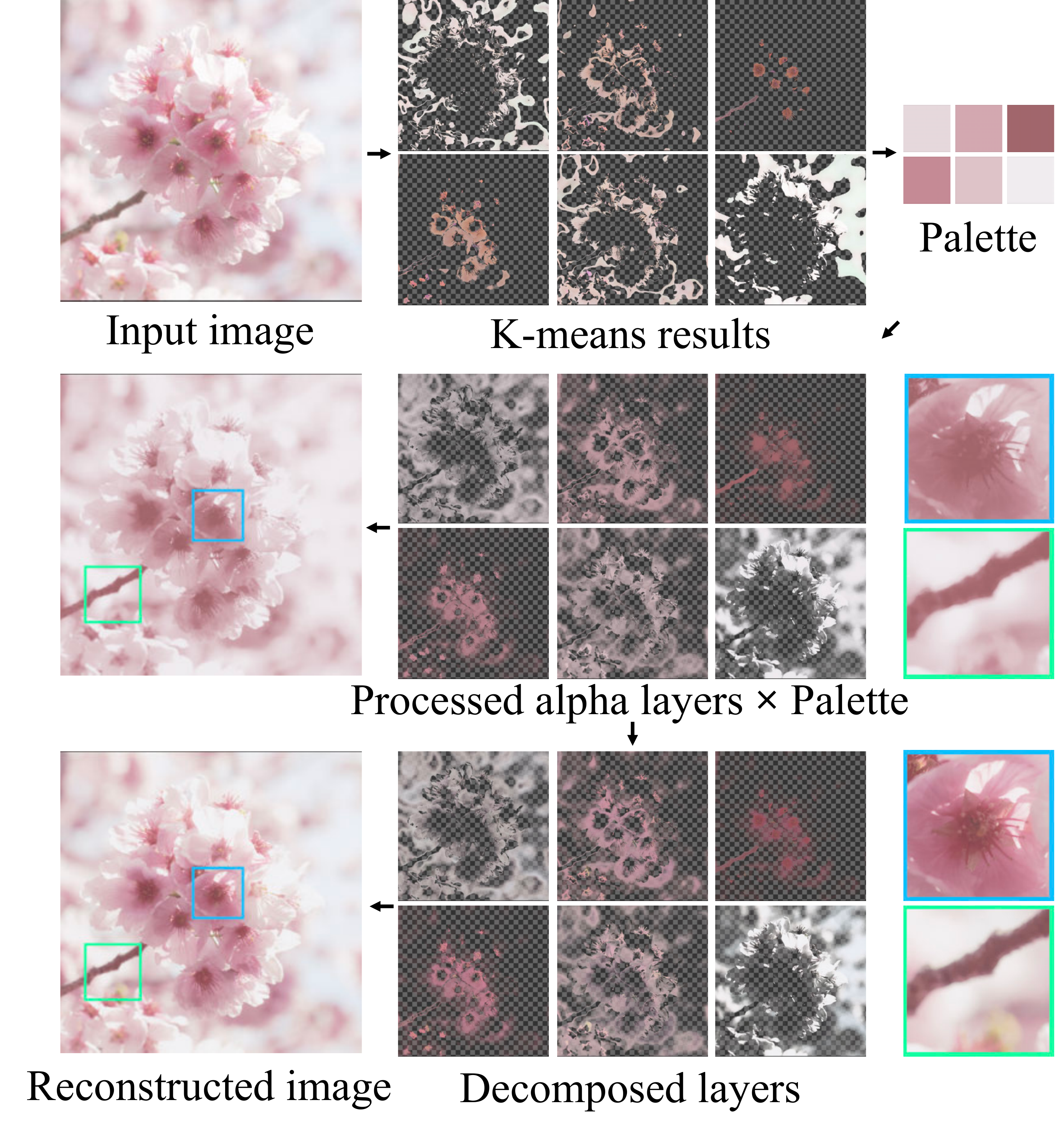}
   \caption{The reconstruction process from Figure \ref{fig:system}. Note that if we do a weighted sum of the palette colors and the corresponding alpha layers, we can also obtain an image (middle left) similar to the input image, but residues are required to recover the input image with various colors and enhanced details (bottom left).}
\label{fig:composition}
\end{center}
\end{figure}

In this work, we propose a neural network based approach that significantly speed up soft color segmentation while retaining the quality of the soft color layers produced.  
Our system consists of three stages, as shown in Figure \ref{fig:system}. 
In 1) \textbf{palette selection},
we automatically or manually select palette colors, each of which indicates the mean color of a target color layer. 
In 2) \textbf{alpha layer estimation}, 
the alpha predictor estimates alpha layers corresponding to the selected palette colors. 
After processing the alpha layers, in 3) \textbf{color layer estimation},  
the residue predictor estimates color residues that indicate the displacement of the colors from the palette color. 
For each pixel in a final RGBA layer, the color value (the RGB part) is the sum of the palette color and its residue at that pixel, and the alpha value (the A part) is taken from the corresponding processed alpha layer.


Compared to existing methods, the proposed method reduces the decomposition time by a factor of 300,000.
We achieve this through optimization of the objective function on the training dataset in advance instead of minimizing an energy function on the input image on the fly, which is common in optimization-base approaches \cite{aksoy16,aksoy17,koyama}.
The training objective of the networks consists of a reconstruction loss, a regularization loss, and a distance loss.
We jointly train the alpha predictor and the residue predictor in a self-supervised manner, and thus no extra manual labeling or annotation are required.
After training, the networks decompose an image in a feed-forward pass.
The speed gain paves the way for practical applications including real-time editing and frame-by-frame video decomposition.

To summarize our main contributions,
\begin{itemize}
  \setlength{\itemsep}{0cm}
  \item We propose the first neural network based approach and a novel training objective for soft color segmentation.
  \item We conduct qualitative and quantitative experiments to demonstrate that our method significantly outpaces state-of-the-art methods while maintaining comparable visual quality.
  \item We apply our method to several practical applications, especially video editing, with unprecedented efficiency.
\end{itemize}

\section{Related Work}
\textbf{Layer decomposition}, actively studied in both computer vision and computer graphics, is the task of decomposing a single image into multiple RGB or RGBA images. For example, reflection removal \cite{fan2017generic, Wen_2019_CVPR, zhang2018single}, haze removal \cite{Gandelsman_2019_CVPR, Ren-ECCV-2016, Zhang_2017_CVPR, 7128396}, and rain removal \cite{8099669, derain_zhang_2018, Li_2019_CVPR} estimate the foreground and background layers and the mixture ratios for each pixel. Segmentation is also one of the layer decomposition tasks. Part segmentation includes human parsing \cite{7053923, Yang_2019_CVPR} or vector graphics layer generation \cite{FLB17, sbai2018vector}. Semantic segmentation has been actively addressed \cite{chen2017deeplab, liu2019auto, long2015fully, ronneberger2015u, zhao2017pyramid}. The segmentation tasks mentioned above classify each pixel into a single class. In contrast to these hard segmentation tasks, {\it soft} segmentation are studied \cite{aksoy16, pan2016soft, sss, tai07} to expresses transitions between regions of classes using alpha value. For instance, matting \cite{chen2013knn, ifm, levin2008spectral, samplenet, wang2018deep, Xu_2017_CVPR, zhang2019late} obtains transitions between a foreground and a background. Soft segmentation is more suitable than hard segmentation for handling motion blur, transparency, illumination, and so on. 

\textbf{Soft color segmentation} is a specific soft segmentation task of decomposing a single image into multiple RGB or RGBA color layers, each of which contains homogeneous colors. The state-of-the-art method proposed by \aksoy \cite{aksoy17} improved the method in \aksoy \cite{aksoy16} by adding a sparsity term to the color unmixing formulation. Research similar to \aksoy \cite{aksoy17} includes \koyama \cite{koyama} and \jtan \cite{tan18}. \koyama \cite{koyama} generalized \aksoy \cite{aksoy17} to enable advanced color blending. \jtan \cite{tan18} are a geometric approach that finds a RGBXY convex hull, extending \jtan \cite{tan16}. Whereas \koyama \cite{koyama} and \jtan \cite{tan16} consider the order of layer stacking, \aksoy \cite{aksoy17} and \jtan \cite{tan18} deal with a {\it linear additive model}, which does not consider the order. \aksoy \cite{aksoy17} have shown that color unmixing based segmentation can be achieved by solving the constrained minimization problem, as explained in Section \ref{section_optimization_approach}. 

\section{Motivation}  \label{section_optimization_approach}

Several works \cite{aksoy17, koyama, tan16} adopt an optimization-based approach to decompose an image into multiple color-considering RGBA layers.
In particular, the state-of-the-art method from \aksoy \cite{aksoy17} proposes the color unmixing formulation, in which the objective function consists of a color constraint, an alpha constraint, a box constraint, and a sparse color unmixing energy function.

To be specific, the color constraint requires that the decomposed layers restore the original image. For a pixel $p$ in the $i^{\rm th}$ layer, 
\begin{equation}
    \sum_i \alpha_i^p \bm{u}_i^p = \bm{c}^p \quad \forall p,
\label{eq:color-constraint}
\end{equation}
where $\bm{c}^p$ denotes an original color at a pixel $p$, and $\bm{u}_i^p$ denotes a layer color at $p$.
We omit the superscript $p$ in the remainder of the paper for convenience.

The alpha constraint enforces the sum of the decomposed layers to be an opaque image:
\begin{equation}
    \sum_i \alpha_i = 1.
\label{eq:alpha-unity}
\end{equation}

The box constraint requires that alpha and color values should be in bound:
\begin{equation}
    \alpha_i, \bm{u}_i \in [0,1] \quad \forall i.
\label{eq:box-constraint}
\end{equation}

The sparse color unmixing energy function is the weighted sum of the distances between colors and the color distribution in each layer, plus a sparsity term:
\begin{equation}
    \mathcal{F}_S = \sum_i \alpha_i \mathcal{D}_i(\bm{u}_i) + \sigma \left( \frac{\sum_i \alpha_i}{\sum_i \alpha_{i}^2} - 1 \right),
\label{eq:energy-function}
\end{equation}
where the layer cost $\mathcal{D}_i(\bm{u}_i)$ is defined as the squared Mahalanobis distance of the layer color $\bm{u}_i$ to the layer distribution $\mathcal{N}(\bm{u}_i, \Sigma_i)$, and $\sigma$ is the sparsity weight.

There is space for improvement of the above approach \cite{aksoy17} in terms of inference speed.
Based on an input image, the iterative optimization of the energy function tends to be slow and running time scales up linearly with the number of pixels, as shown in Section \ref{section_quantitative_evaluation}. 

The success of optimzation-based methods \cite{aksoy17, koyama, tan16} and its disadvantage in speed inspire us to train neural networks on the dataset.
Without any on-the-fly iteration, the networks decompose the original image into soft color layers in a significant lower inference time.
Partly inspired by the minimization of the energy function, we come up with an objective function that is designed for training our neural networks system, as detailed in the next section. 

\section{Methods} \label{section_methods}
Our proposed method consists of three stages: palette color selection, alpha layer estimation and color layer estimation. The input of the proposed system is a single RGB image, and the outputs are RGBA soft color layers.

\subsection{Palette Color Selection}
The inputs of the methods of both \aksoy and \koyama \cite{aksoy17, koyama} are a set of color models that represent the means and the (co)variances of colors of the desired output layers. 
Although covariances provide additional controllability, a user has to understand the deﬁnition of the covariance of colors, how to adjust it, and how it interacts with the system to produce the final color layers. 
Consequently, the user may not find it intuitive to manipulate covariances.
Aiming for an easy-to-use user experience, we believe that a user should not be exposed to more parameters than necessary.
Therefore, we make the design choice to take palette colors (means only) instead of color models (means + covariances) as inputs.
A palette color has the simple interpretation as the mean value of the colors that should be included in a color layer. 

During training, we use K-means algorithm to partition pixels in an input image into $K$ clusters in the 3-dimensional RGB space.
We pick the center RGB values of the clusters as the palette colors. 
The number of palette colors $K$ is fixed throughout the training of the networks. 
During inference, the palette colors can be specified manually, in addition to automatic selection using K-means.

\subsection{Alpha Layer Estimation}
We adopt the U-Net architecture \cite{ronneberger2015u} for our {\it alpha predictor}. 
The inputs of the alpha predictor are the original RGB image and $K$ single-color images.
Each of the single-color image is simply an RGB image filled with a single palette color. 
The outputs are $K$ alpha layers, which are single-channel images of alpha values.
In the subsequent {\it alpha layer processing} in Figure \ref{fig:system}, the outputs of the network are normalized by:
\begin{equation}
    \alpha_i = \frac{\alpha_i}{\sum_k \alpha_k},
\label{eq:normalization}
\end{equation}
where $\alpha_i$ is the alpha value (opacity) at a certain pixel position. 
This normalization step ensures that the output satisfies the alpha-add condition (Eq. (\ref{eq:alpha-unity})).
For inference, we can add various kinds of alpha layer processing in addition to normalization, which we detail in Section 4.4.

\subsection{Color Layer Estimation}
The palette colors and the alpha layers are not sufficient to reconstruct the original image, as shown in Figure (\ref{fig:composition}). 
Although a user only needs to specify the palette colors, each color layer should contains a more variety of colors than a single palette color.

To introduce color variations, we add a {\it residue predictor} to estimate color residues from the palette colors. 
The inputs of the residue predictor are the original image, $K$ single-color images, and $K$ normalized alpha layers. 
The residue predictor network has an identical architecture as the alpha predictor, except for the number of input and output channels. 
We add the output residues to palette colors to compute $K$ RGB layers. 
The final RGBA layers can be obtained by concatenating RGB layers and normalized alpha layers along the channel axis.

\begin{figure*}[t]
\begin{center}
\includegraphics[width=\linewidth]{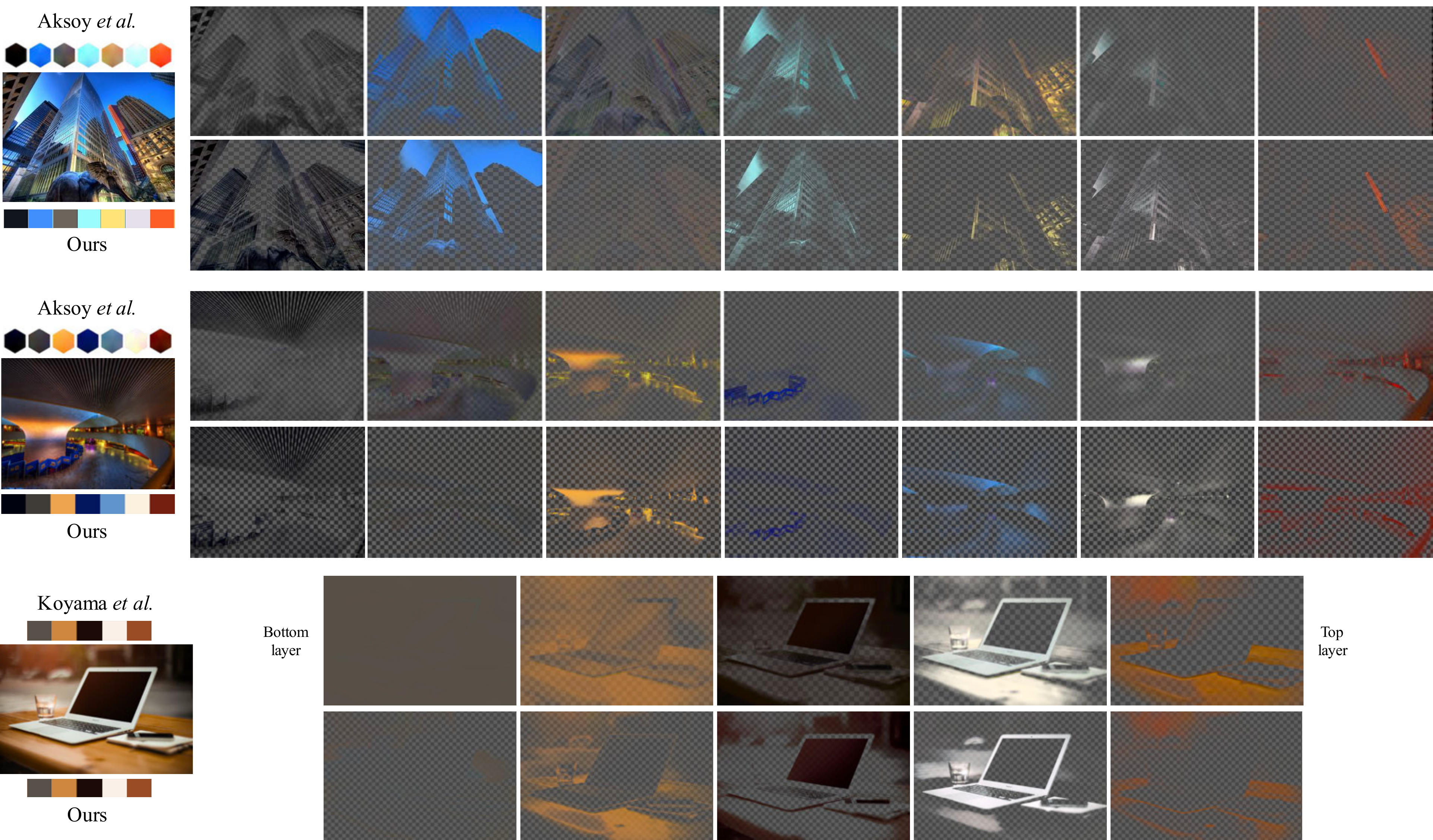}
   \caption{Qualitative comparison with the state-of-the-art baselines \cite{aksoy17, koyama}. Our method performs comparably to the state-of-the-art methods \cite{aksoy17} in producing meaningful soft color layers. We note that Koyama {\it et al.} \cite{koyama} deals with order-dependent blending while we are order-independent.
   }

\label{fig:qualitative}
\end{center}
\end{figure*}


\subsection{Network Training}
The above two networks are trained end-to-end with an objective composed of several losses. The main training objective, inspired by the color constraint (Eq. (\ref{eq:color-constraint})), is to minimize a self-supervised reconstruction loss between the input and the output:
\begin{equation}
    L_r = \| \sum_i \alpha_i \bm{u_i} - \bm{c} \|_1.
\label{eq:r_loss}
\end{equation}

To regularize the training of the alpha predictor, we propose a novel loss for regularization, formulated as the reconstruction loss between the original image and the reconstructed image {\it without} color residues:
\begin{equation}\label{eq:regularization_loss}
    L_a = \| \sum_i \alpha_i \bm{p_i} - \bm{c} \|_1,
\end{equation}
where $\bm{p_i}$ denotes the palette color of the $i^{\rm th}$ layer. In other words, $\sum_i \alpha_i \bm{p_i}$ is an image that is reconstructed only using the normalized alpha layers and the palette colors, as shown in the middle left of Figure (\ref{fig:composition}).

To gather only homogeneous colors in each color layer, we propose a novel distance loss, in reminiscence of Eq. (\ref{eq:energy-function}), formulated as 
\begin{equation}\label{eq:distance_loss}
    L_d = \sum_i \alpha_i \| \bm{p_i} - \bm{u_i} \|_2.
\end{equation}
We use Euclidean distance in RGB space because our inputs are simply $K$ palette colors. 

Now we are ready to formulate our total loss as follows:
\begin{equation}
    L_{total} = L_r + \lambda_a L_a + \lambda_d L_d,
\label{eq:total-loss}
\end{equation}
where $\lambda_a$ and $\lambda_d$ are coefficients for regularization loss and distance loss, respectively.
In comparison with the method of \aksoy \cite{aksoy17} which minimizes the proposed energy function for a given input image, our method trains neural networks to minimize the total loss on a training dataset.

We note that 1) the outputs of the networks automatically satisfy the box constraint (Eq. (\ref{eq:box-constraint})) because of the sigmoid functions and clip operations appended to the networks. And 2) we do not enforce sparsity like the sparsity term in sparse color unmixing energy function (Eq. (\ref{eq:energy-function})). In preliminary experiments we introduced such a sparsity loss with a coefficient to control its weight in the total loss. When its weight is high, there are no soft transition on the boundary of regions, resulting in nearly hard segmentation. When we decrease the weight, however, the sparsity loss no longer promotes sparsity. We believe that the novel regularization loss $L_a$ and the normalization step in alpha layer processing (Eq. (\ref{eq:normalization})) have collectively encouraged the sparsity of alpha layers, and therefore it is redundant to introduce an extra sparsity loss.


\section{Experiments}
For training, we use Places365-Standard validation images \cite{places}. All images in the training dataset have a resolution of $256 \times 256$. We use the Adam optimizer with $lr = 0.2$, $\beta_1 = 0.0$, and $\beta_2 = 0.99$. For all experiments, we set $\lambda_a = 1$ and $\lambda_d = 0.5$.

For inference, since our networks are fully-convolutional, we can apply the model to decompose images of various resolutions. Both of the networks are U-Net with five convolutions, three transposed convolutions, and skip connections at each scale. We discuss the network structures in detail in the supplementary material.

\subsection{Qualitative Evaluation}

\textbf{Comparison with state-of-the-art methods.} Figure \ref{fig:qualitative} shows a qualitative comparison between our results and results of \aksoy \cite{aksoy17} or \koyama \cite{koyama}. Note that the inputs are different: their inputs are distributions of colors, while ours are simply colors. 

We would like to mention that there is no optimal solution for soft color segmentation, because an image can be decomposed into meaningful and high-quality color layers in various way. As an attempt to make the comparison as intuitive as possible, we manually select palette colors for our method so that the result color layers look similar to the result in \aksoy and \koyama \cite{aksoy17, koyama}.

\subsection{Quantitative Evaluation} \label{section_quantitative_evaluation}
\begin{figure}[t]
\begin{center}
\includegraphics[width=0.8\linewidth]{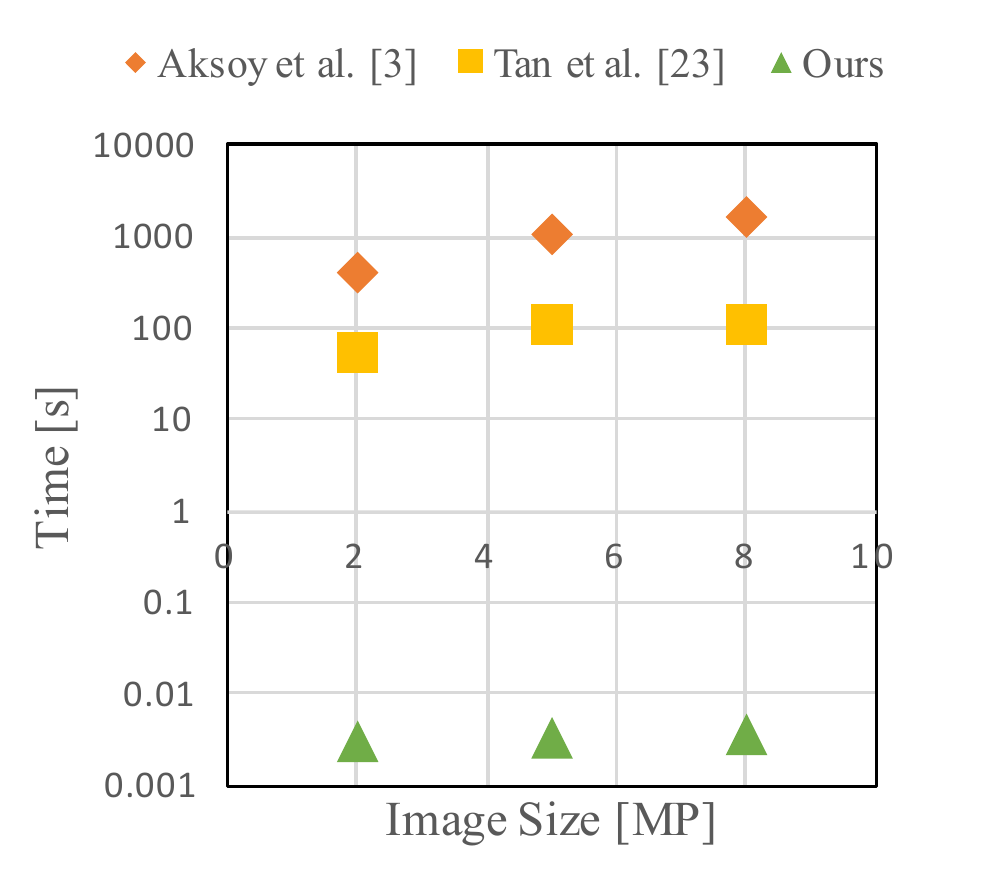}
   \caption{Speed comparison with the state-of-the-art methods. Note that the time axis is in logarithmic scale. 
   }
\label{fig:speed}
\end{center}
\end{figure}

\textbf{Speed.} Figure \ref{fig:speed} shows the running time of our algorithm,  algorithm of \aksoy and \jtan \cite{aksoy17, tan18}, depending on the size of the input image.
Both \aksoy \cite{aksoy17} and our method use a palette size of 7, and \jtan \cite{tan18} use an average palette size of 6.95, and the median palette size is 7.
It takes 2.9 ms for the proposed method to decompose a 1080p resolution (about 2 MP) image, 3.0 ms for a 5 MP image and 3.4 ms for a 4K resolution (about 8 MP) image, averaged from 20 experiments at each resolution. \aksoy  \cite{aksoy17} reports that their algorithm takes 1,000 s to decompose a 5 MP image. Their running time scales up linearly as the number of pixels grows due to the per-pixel optimization. \jtan \cite{tan18} improve the execution speed of \aksoy, utilizing RGBXY convex hull computation and layer updating. We collect their running time (about 50 s to 100 s) in Figure 9 in their paper \cite{tan18} because exact values are not reported. See supplementary material for detail experimental settings.

The comparison shows that our neural network based method has a significant speed improvement over the state-of-the-art methods.
Furthermore, only our method can decompose a video in a practical amount of time. Specifically, when the video consists of 450 frames (1080p, 30 fps, 15 s), our method takes 1.35 seconds, while the method of \aksoy  \cite{aksoy17} takes around 50 hours.

\textbf{Reconstruction error.} 
In Table \ref{table:quantitative-comparison}, We use pixel-level mean squared error to evaluate the difference between an input image and the reconstructed image.
Althought we minimize the reconstruction loss (Eq. (\ref{eq:r_loss})) on a training dataset instead of the input image as \aksoy \cite{aksoy17} do, 
our reconstruction errors are sufficiently low, indicating that our networks generalize well on the test images. 
We note that the reconstruction error also depends on the palette colors selected. In particular, the error increases if some palette colors never appear in the original image at all.

\begin{table}
\footnotesize
\begin{center}
\begin{tabular}{lcccc}
\toprule
Method & Reconst. $\downarrow$ & PSNR $\uparrow$ & SSIM $\uparrow$ & Sparsity $\downarrow$ \\
\midrule
Aksoy {\it et al.} \cite{aksoy17} & {\bf 0.00050} & - & - & - \\
Ours & 0.00088 & 31.07 & 0.9740 & 1.456\\
\midrule
w/o $L_r$ & 0.00308 & 25.70 & 0.9158 & 1.279 \\
w/o $L_a$ & 0.00090 & 31.17 & {\bf 0.9750} & 1.959 \\
w/o $L_d$ & 0.00076 & 31.72 & 0.9743 & 1.640 \\
w/o Skip & 0.00350 & 27.37 & 0.9366 & {\bf 1.149} \\
w/o Zero-centered & 0.00073 & {\bf 31.82} & 0.9710 & 1.450 \\
Single network & 0.00104 & 30.71 & 0.9633 & NaN \\
\bottomrule
\end{tabular}
\caption{Quantitative comparison. Our method is the only setting in the ablation study that achieves similar reconstruction error as \aksoy \cite{aksoy17} with both low sparsity score and high image quality. See \ref{section_quantitative_evaluation} for details of the quantitative evaluation, and \ref{section_ablation_study} for details of the ablation study. }
\label{table:quantitative-comparison}
\end{center}
\end{table}

\begin{figure*}[t]
\begin{center}
\includegraphics[width=0.8\linewidth]{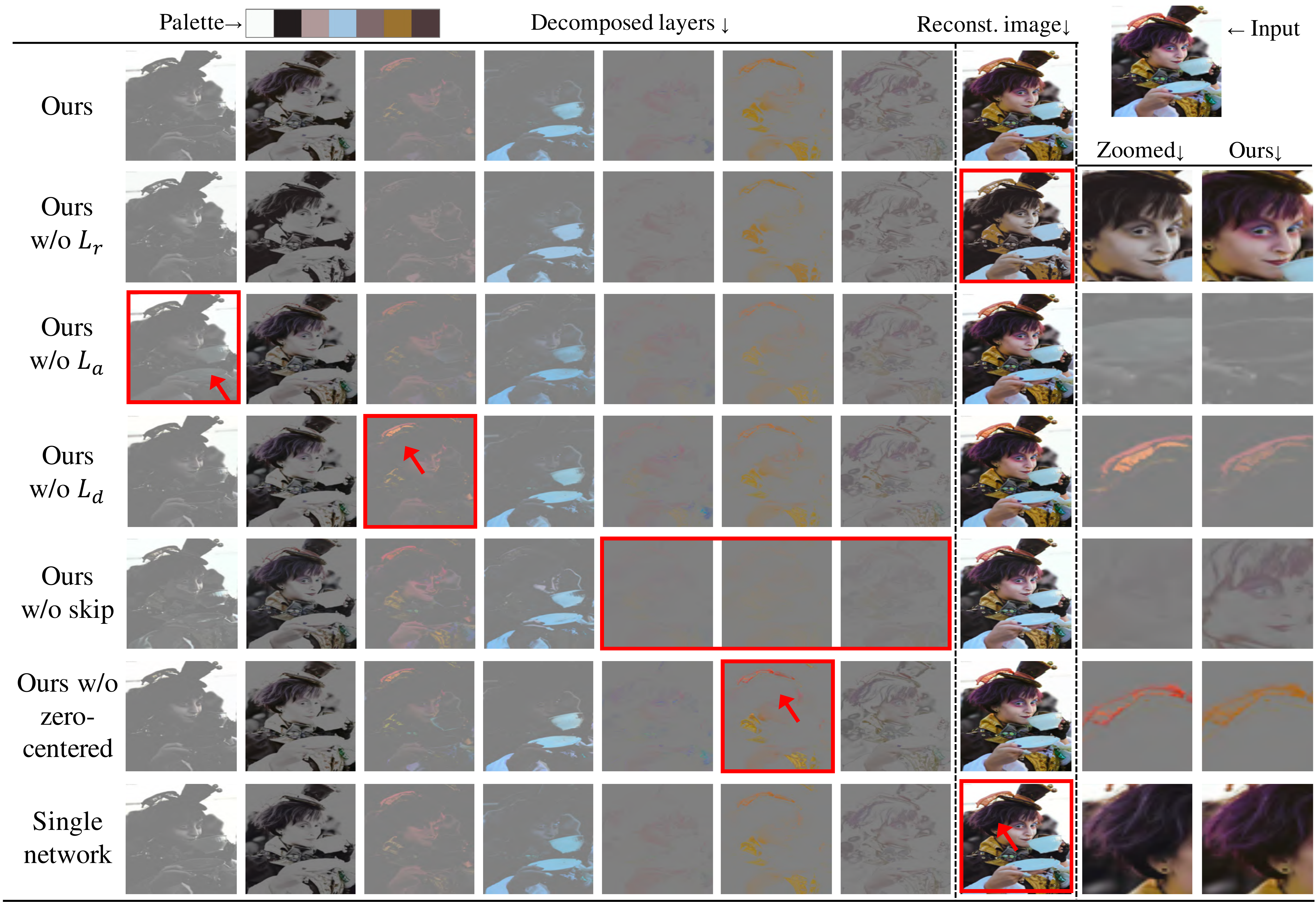}
   \caption{Visual comparison for ablation study. Removing any component of our method results in higher reconstruction error, color contamination or unnecessary overlapping of color layers. See section \ref{section_ablation_study} for details. }
\label{fig:ablation-quality}
\end{center}
\end{figure*}

\begin{table*}
\small
\begin{center}
\begin{tabular}{lccccc}
\toprule
Method & Aksoy {\it et al.} \cite{aksoy17} & Ours & w/o $L_a$ & w/o $L_d$ & w/o Zero-centered \\
\midrule
Color Var. $\downarrow$ & 0.005 &  {\bf 0.003} & 0.007 & 0.006 & 0.162 \\
\bottomrule
\end{tabular}
\caption{Quantitative comparison of color variance.
The values of our methods correspond to the images in Figure \ref{fig:ablation-quality}.
We obtain the color variance of the method of \aksoy from their paper \cite{aksoy17}.
}
\label{table:color-var}
\end{center}
\end{table*}

\subsection{Ablation Study} \label{section_ablation_study}
In this section, we validate the losses and architectures of the neural networks. Figure \ref{fig:ablation-quality} shows a sample for qualitative comparison.
The reconstruction error and the sparsity scores shown in Table \ref{table:quantitative-comparison} are the averaged scores of 100 images of 1 MP or more. The sparsity score is calculated as
\begin{equation}
    L_s = \frac{\sum_i \alpha_i}{\sum_i \alpha_{i}^2} - 1.
\end{equation}
A lower value of $L_s$ means the decomposed layers are sparser.
The color variance score in Table \ref{table:color-var} represents the score corresponding to the results in Figure \ref{fig:ablation-quality}.
This score is the sum of individual variances of the RGB channel averaged over all decomposed layers. For fair comparison, we used same palette colors to decompose an input for each setting.

\textbf{Ours versus ours without $\bm{L_r}$.}
Table \ref{table:quantitative-comparison} and Figure \ref{fig:ablation-quality} show that our method without $L_r$ cannot properly reconstruct an input image. $L_a$ trains the only {\it alpha predictor}, so the {\it residue predictor} cannot function properly.

\textbf{Ours versus ours without $\bm{L_a}$.}
Although the SSIM score is marginally better without $\bm{L_a}$, the sparsity score is significantly higher, which suggests that excessive overlapping exists between alpha layers.
On the third row in Figure \ref{fig:ablation-quality}, the blue plate on the person's hand wrongly shows up in the white layer (highlighted in red), causing overlapping between white and blue layers.
Overlapping is intended to occur only sparsely, e.g. at the boundary of a region, because excessive overlapping is not suitable for application to image editing.

Moreover, the color variance score is higher without $\bm{L_a}$, indicating that some of the layer might be contaminated, i.e. containing colors that is much different from the corresponding palette color.
We can observe such contamination on the top-left corner of the blue layer.
We believe this is because $L_a$ improves the performance alpha predictor.


\textbf{Ours versus ours without $\bm{L_d}$.}
Without $\bm{L_d}$ suppressing the variance of colors in each layer, the reconstruction error decreases. However, a large color variance causes the same color to spread across multiple layers, which is not desirable for color-based editing, as shown in Table \ref{table:color-var} and Figure \ref{fig:ablation-quality}.


\textbf{Ours versus ours without skip connection.}
Without skip connections in the {\it alpha predictor} and the {\it residue predictor}, alpha layers are not accurate, leading to a higher the reconstruction error.

\textbf{Ours versus ours without zero-centered residues.}
It is easier to train a neural network with zero-centered output, and we can make sure the palette color is the mean value of each layer. Without zero-centering, the PSNR increases, but the color variance increases even more. 

\textbf{Ours versus a single network only.}
If we use a plain single network, the reconstruction error increases. 
We believe that accurate alpha layers as inputs enhance the performance of the residue predictor.
Specifically, in our method, we apply smoothing ﬁlters to remove checkerboard artifacts from alpha layers, as shown in Figure \ref{fig:same-color}, and predict the RGB channels based on the processed alpha layers. 
We doubt that there is a way to incorporate smoothing ﬁlter processing in a neural network that predicts both alpha and RGB channels simultaneously.

\begin{figure}[t]
\begin{center}
\includegraphics[width=\linewidth]{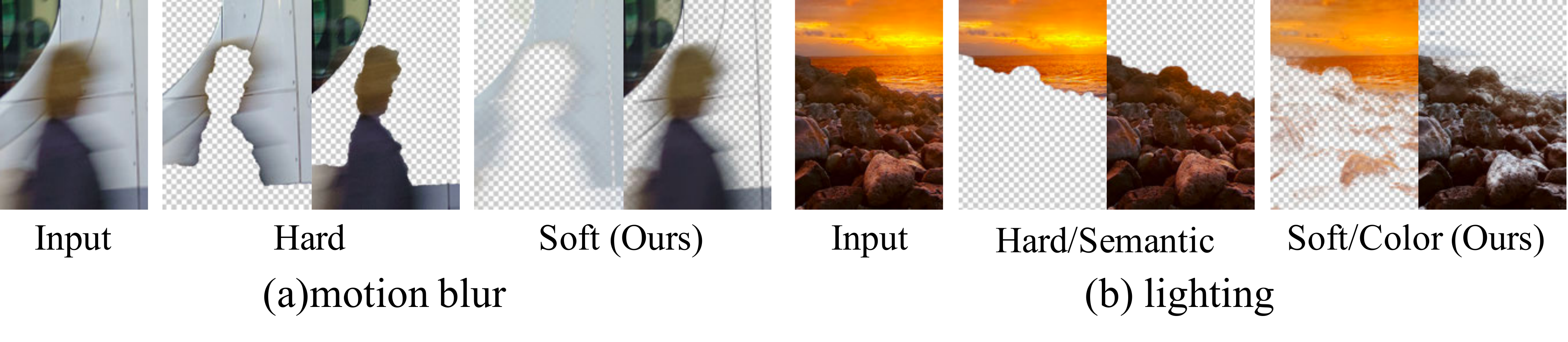}
   \caption{
   Comparison with hard segmentation and color-based soft segmentation on images with ambiguous object boundaries.
   In (a), hard segmentation produces a background layer that is tainted by the moving object in the foreground.
   In (b), hard semantic segmentation divides the image into semantic areas, but soft color segmentation enable us to change simultaneously the reflected light on the rock along with the lighting condition of the sky.
   }
\label{fig:usefulcase}
\end{center}

\end{figure}

\subsection{Applications}

\textbf{Decomposition of images with blurry boundary.}
Soft color segmentation is useful not only for decomposing an image into soft color layers, but for any case where the resulting layers are preferred to have blurry boundary. 
Figure \ref{fig:usefulcase} shows the results of our decomposition and the benefits of soft color segmentation.
In the case of an image with a foreground object with motion blur, although both the foreground and the background are visible at a single pixel, hard segmentation has to assign that pixel into either of the classes.
In such a case, soft segmentation can recognize the pixel as a mixture of fore/background and thus has an advantage over conventional hard segmentation.

\begin{figure}[t]
\begin{center}
\includegraphics[width=0.95\linewidth]{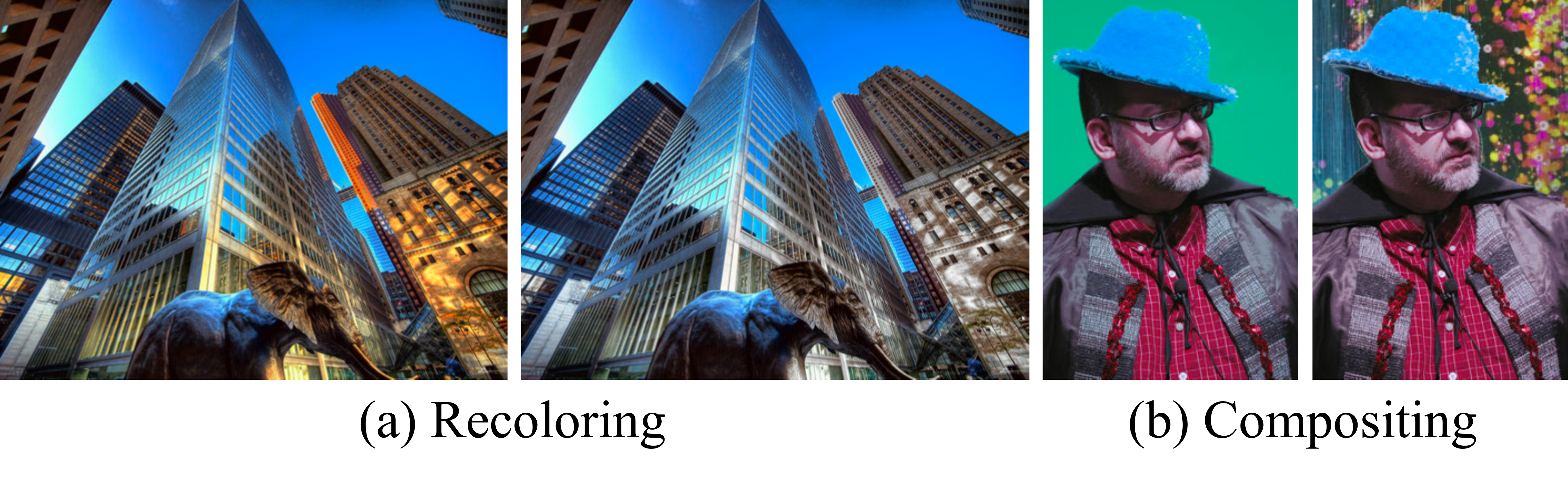}
   \caption{Examples of image editing using the layers decomposed by our method.}
\label{fig:image-editing}
\end{center}

\begin{center}
\includegraphics[width=0.95\linewidth]{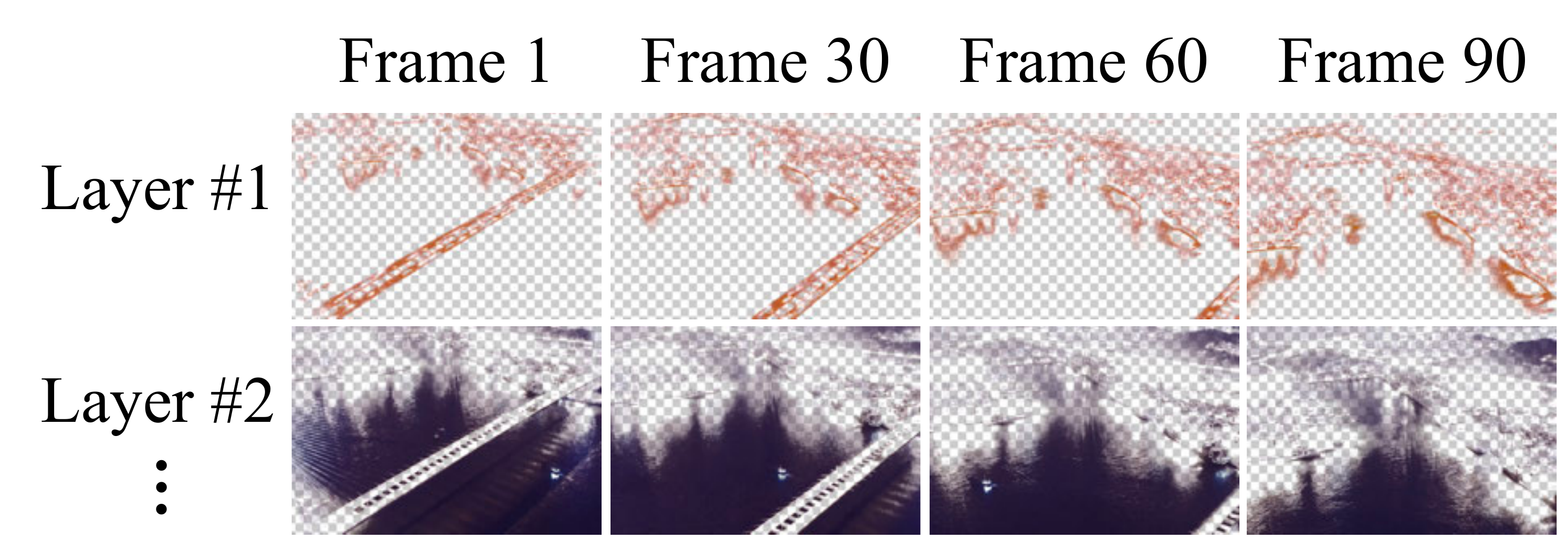}
   \caption{Examples of video decomposition. Our method decomposes all the frames in the video in a few seconds.}
\label{fig:video-decomposition}
\end{center}

\begin{center}
\includegraphics[width=0.95\linewidth]{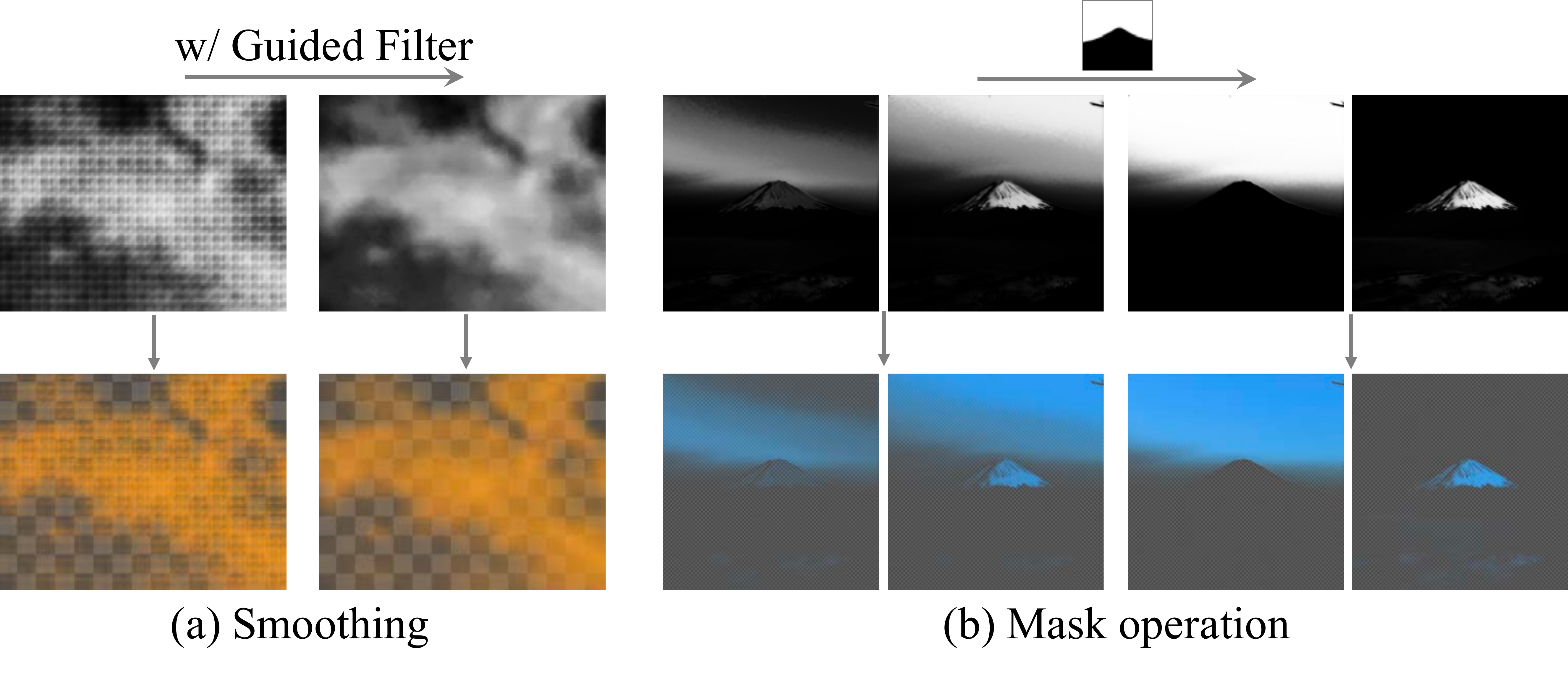}
   \caption{Examples of alpha layer processing before predicting colors. (a) Comparison between with/without smooth filtering. (b) An example of using a mask to manipulate the decomposed layers.}
\label{fig:alpha-processing}
\end{center}

\begin{center}
\includegraphics[width=0.95\linewidth]{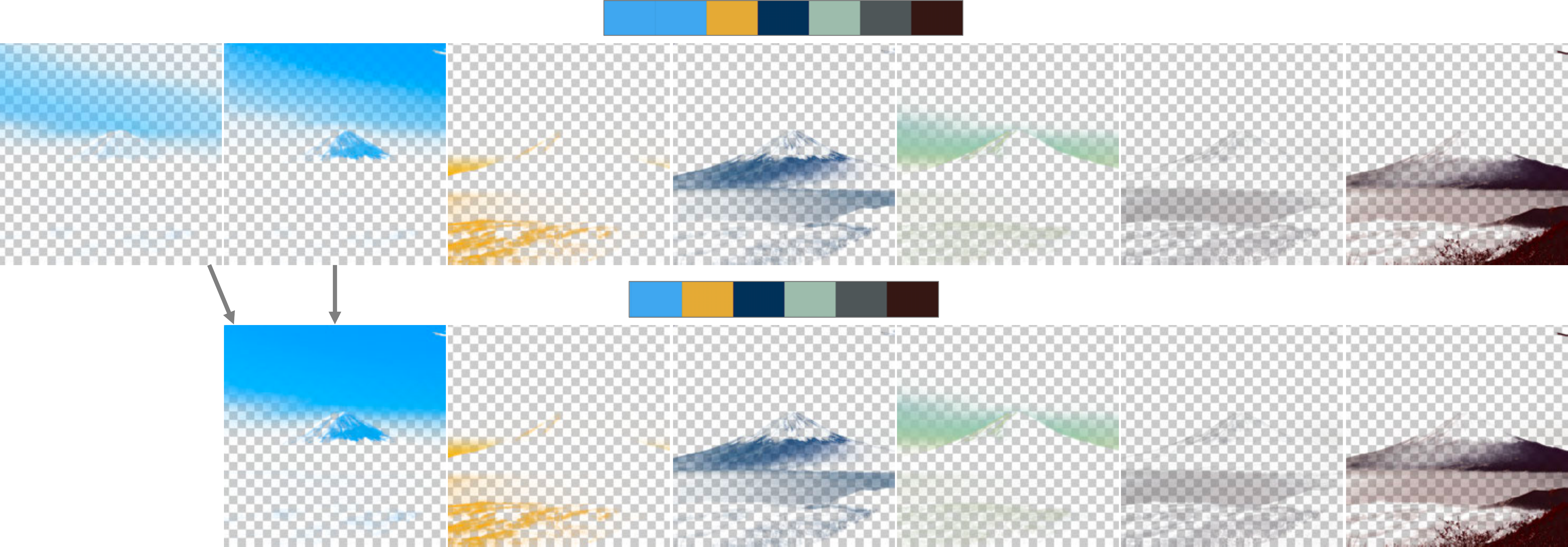}
   \caption{An example when the palette contains a duplicated color. It shows that, although the number of layers is fixed during training, our model tolerates the duplication of colors and is thus capable of decomposing an image into fewer layers.}
\label{fig:same-color}
\end{center}
\end{figure}

\textbf{Natural image editing.}
Figure \ref{fig:teaser} and Figure \ref{fig:image-editing} show examples of recoloring and compositing. These editing results are created by editing each decomposed layer with the {\it alpha add} mode in Adobe After Effects.

\textbf{Video decomposition.}
Figure \ref{fig:video-decomposition} shows our method performing video soft color segmentation. We decompose the video frame-by-frame, without any constraints for temporal consistency. 
Nevertheless, the decomposed layers do not flicker. 
It can be partly attributed to applying smoothing ﬁlter and ﬁxing color palettes, both of which encourage consistent alpha layers, and regularization loss $\bm{L_d}$, which encourages consistent color layers.
Compared to other methods, only our method can decompose video in practical time. 
See the video material for detailed results.

\textbf{Alpha layer processing.}
In the {\it alpha layer processing} stage, a user can edit the predicted alpha layers, and subsequently use these edited alpha layers for color estimation, thanks to the fact that the estimation of alpha layers is independent from color layers. 
As shown in Figure \ref{fig:alpha-processing}, we can use the guided filter \cite{gfilter, fastgfilter} to smooth the image, or manipulate a mask to change the alpha region. 
We can prepare the mask manually or automatically, capitalizing on state-of-the-art methods (e.g. semantic segmentation and depth estimation). 
Therefore, our method can be complementary to various image editing techniques.


\subsection{Limitations}
\textbf{Memory limit.} 
Because we use a GPU for inference, we cannot handle high-resolution images that exceed the GPU memory limit. Also, sufficient GPU memory is needed to hold the intermediate features of the encoder-decoder networks with skip connections. Concretely, when an input is 1080p (4K) resolution and computation is based on 32-bit floating point, about 12 (21) GB is consumed. Countermeasures include using 16-bit floating point and discarding unnecessary intermediate features. 

\textbf{Fixed palette size.} In our approach, the number of decomposed layers for each trained model is fixed. To handle various numbers of layers, one solution is to train a model that decomposes an image into the sufficient number of layers, and use palettes with duplicated colors, as shown in Figure \ref{fig:same-color}. We merge layers of duplicated colors after decomposition.

\section{Conclusion}
In this paper, we tackled the problem of soft color segmentation using the first neural network based approach. We showed that the proposed method enabled faster decomposition relative to prior state-of-the-art methods. We also demonstrated that our fast and simple approach enabled new important applications, such as soft color segmentation of videos.

\section*{Acknowledgement}
We are grateful for Yingtao Tian and Prabhat Nagarajan for helpful advice on writing the paper. We thank Edgar Simo-Serra and all reviewers for helpful discussions and comments. 
{\small
\bibliographystyle{ieee_fullname}
\bibliography{egbib}
}

\newpage
\appendix
\appendixpage

\section{Network Structures}
In Table \ref{table:alpha-predictor} and \ref{table:residue-predictor}, we show the  architectures of the {\it alpha predictor} and the {\it residue predictor}, based on the naming conventions of network components below:
\begin{itemize}
  \item Conv2d(K, P): 2D convolution with the kernel size of K and the padding of P;
  \item DeConv2d(K, P): 2D transposed convolution with the kernel size of K and the padding of P;
  \item BN: Batch normalization.
\end{itemize}

\section{Experimental Settings of Speed Test}
When we compare decomposition speed of methods of \aksoy \cite{aksoy17}, \jtan \cite{tan18} and ours, \aksoy use a palette size of 7, and \jtan use an palette size with a mean of 6.95 and median of 7. The methods of measurement for each algorithm are presented below. 
\begin{itemize}
    \item We measure the running time as the total time of alpha layer estimation and color layer estimation (not including reading an input image into GPU memory).
    We execute our Python code on a 3.50 GHz Intel Core i7-7800X CPU and 64GB of RAM and a NVIDIA Quadro P6000 GPU.  
    At each resolution, we average the decomposition time over 20 images as the final results.
    \item \aksoy \cite{aksoy17} use parallelized C++ to conduct the experiment. 
    \item 
    The running time reported by \jtan \cite{tan18} including the execution time of RGBXY convex hull computation and layer updating.
    They execute their Python code on a 2.9 GHz Intel Core i5-5257U CPU and 16 GB of RAM. 
    In their method, a layer updating is possible in few milliseconds, but there is no way to bypass the intensive computation of RGBXY convex hull for each image.
\end{itemize}

\section{Qualitative Comparisons}
\begin{figure*}[t]
\begin{center}
\includegraphics[width=1.0\linewidth]{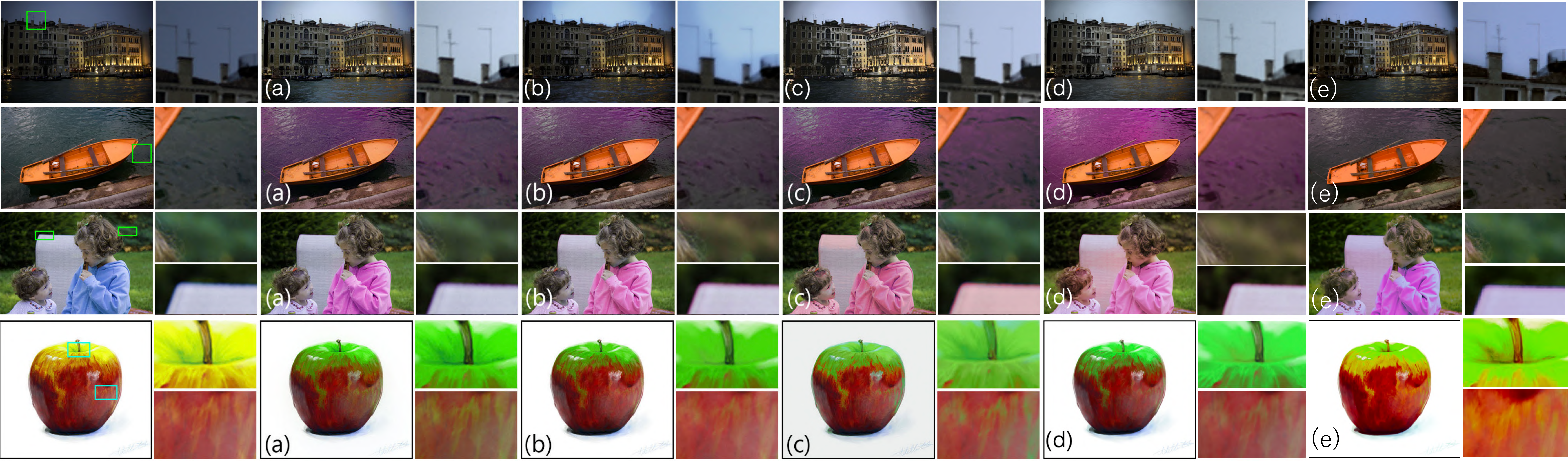}
   \caption{Comparisons between previous approaches and our algorithm on recoloring. From left to right: (a) Aksoy {\it et al.} \cite{aksoy17}, (b) Tan {\it et al.} \cite{tan16}, (c) Chang {\it et al.} \cite{chang2015palette}, (d) Tan {\it et al.} \cite{tan18} and (e) our approach. This figure extends the comparisons of recoloring from Tan {\it et al.} \cite{tan18}.}
\label{fig:supp-comp-edit}
\end{center}
\end{figure*}

\begin{figure*}[t]
\begin{center}
\includegraphics[width=1.0\linewidth]{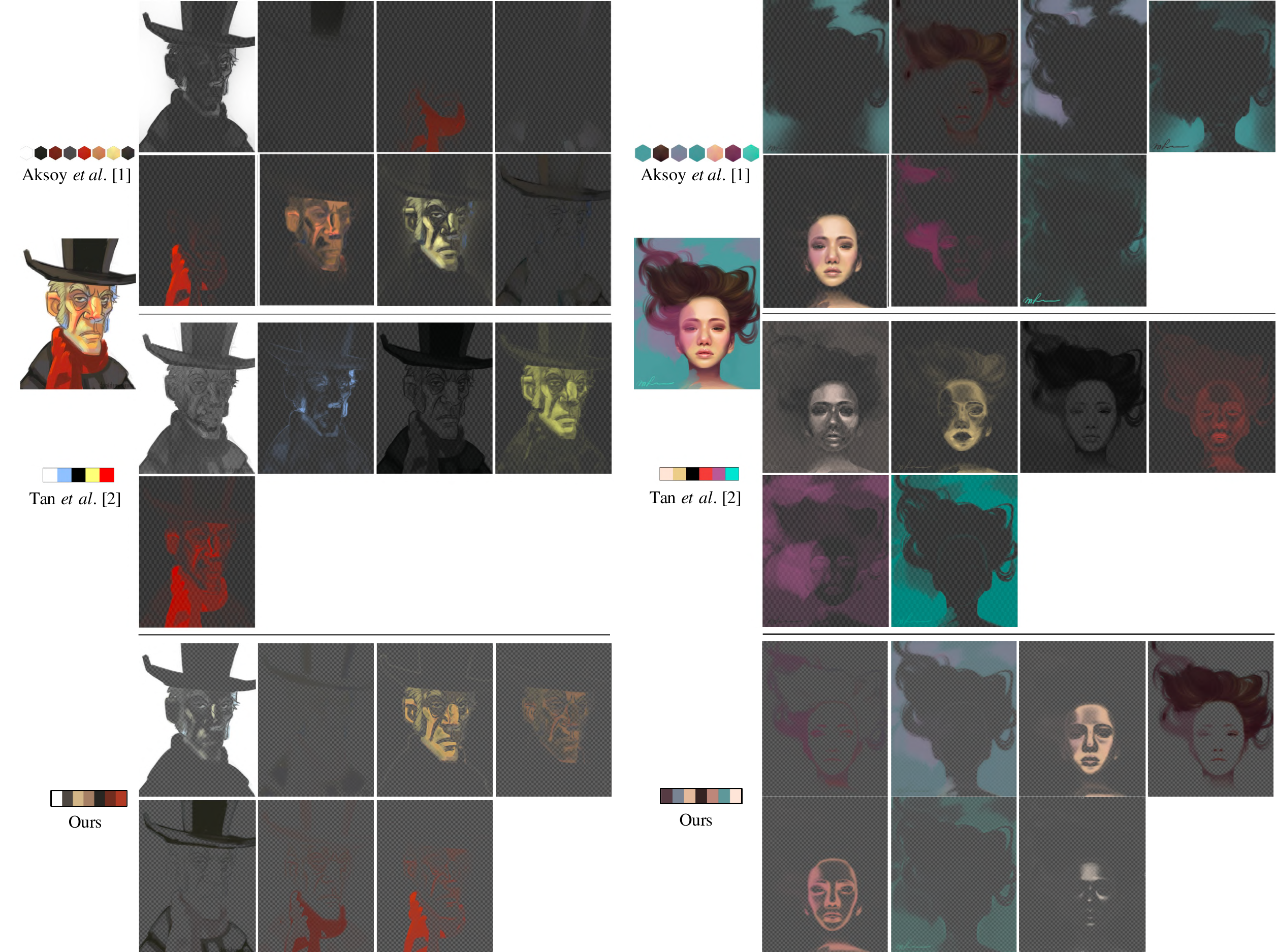}
   \caption{Comparisons of decomposed layers between Aksoy {\it et al.} \cite{aksoy17}, Tan {\it et al.} \cite{tan18}, and our approach.}
\label{fig:supp-comp-represent}
\end{center}
\end{figure*}

\begin{figure*}[t]
\begin{center}
\includegraphics[width=0.84\linewidth]{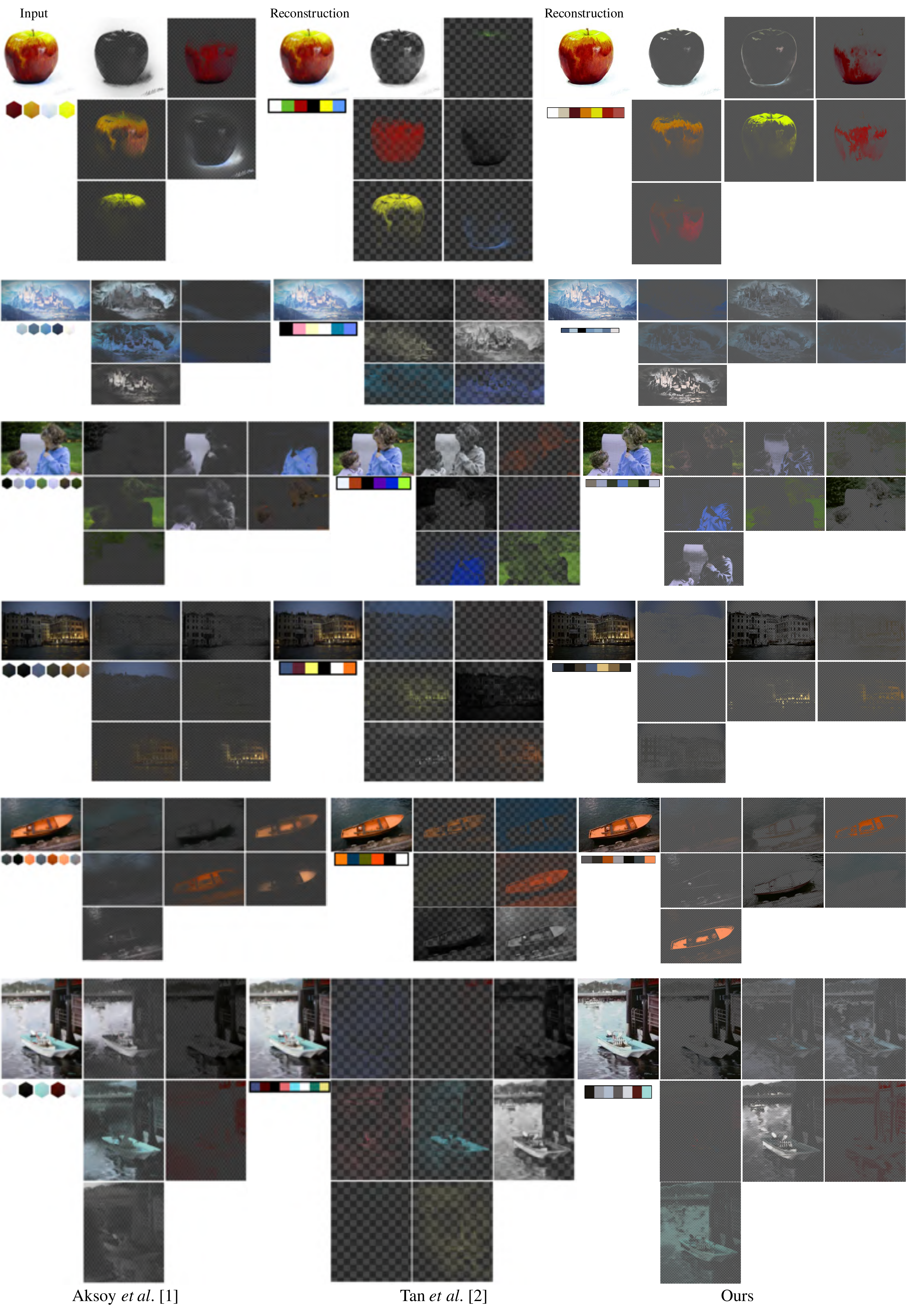}
   \caption{More comparisons of decomposed layers between Aksoy {\it et al.} \cite{aksoy17}, Tan {\it et al.} \cite{tan18}, and our approach.}
\label{fig:supp-comp-layers}
\end{center}
\end{figure*}

To qualitatively evaluate our method, we compare previous methods with ours on recoloring and decomposition.
Figure \ref{fig:supp-comp-edit} shows examples of recoloring, and Figure \ref{fig:supp-comp-represent} and \ref{fig:supp-comp-layers} compare decomposed layers of our method with those of Aksoy {\it et al.} \cite{aksoy17} and Tan {\it et al.} \cite{tan18}.

\begin{table*}
\small
\begin{center}
\begin{tabular}{lccc}
\toprule
Components & Input size & Output size & Output name \\
\midrule
Conv2d(3,1), ReLU, BN & H $\times$ W $\times$ C &  (H/2) $\times$ (W/2) $\times$ (C$\times$2) & Conv-1\\
Conv2d(3,1), ReLU, BN & (H/2) $\times$ (W/2) $\times$ (C$\times$2) &  (H/4) $\times$ (W/4) $\times$ (C$\times$4) & Conv-2\\
Conv2d(3,1), ReLU, BN & (H/4) $\times$ (W/4) $\times$ (C$\times$4) &  (H/8) $\times$ (W/8) $\times$ (C$\times$8) & - \\
DeConv2d(3,1), ReLU, BN & (H/8) $\times$ (W/8) $\times$ (C$\times$8) &  (H/4) $\times$ (W/4) $\times$ (C$\times$4) & Deconv-1\\
Concatenate(Deconv-1, Conv-2) & - & (H/4) $\times$ (W/4) $\times$ (C$\times$8) & - \\
DeConv2d(3,1), ReLU, BN & (H/4) $\times$ (W/4) $\times$ (C$\times$8) &  (H/2) $\times$ (W/2)$\times$ (C$\times$2) & Deconv-2\\
Concatenate(Deconv-2, Conv-1) & - & (H/4) $\times$ (W/4) $\times$ (C$\times$4) & - \\
DeConv2d(3,1), ReLU, BN & (H/2) $\times$ (W/2) $\times$ (C$\times$4) & H $\times$ W $\times$ (C$\times$2) & Deconv-3\\
Concatenate(Deconv-3, Input image) & - & (H/4) $\times$ (W/4) $\times$ (C$\times$2+3) & - \\
Conv2d(3,1), ReLU, BN & H $\times$ W $\times$ (C$\times$2+3) &  H $\times$ W $\times$ C & - \\
Conv2d(3,1), Sigmoid & H $\times$ W $\times$ C & H $\times$ W $\times$ ${\rm C_{out}}$ & - \\
\bottomrule
\end{tabular}
\caption{The network architecture of the alpha predictor that estimates alpha layers from an input image. Specifically, to predict $7$ alpha layers, the alpha predictor takes as inputs an image and $7$ palette layers of size $H \times W \times 3$ ($1 \times 1 \times 3$ palette colors broadcast across spatial dimensions). Therefore, the total number of input channels is $C = 3 + 7 \times 3$. The output, composed of $7$ single-channel alpha layers, has ${\rm C_{out}} = 7$ channels.}
\label{table:alpha-predictor}
\end{center}
\end{table*}

\begin{table*}
\small
\begin{center}
\begin{tabular}{lccccc}
\toprule
Components & Input size & Output size & Output name \\
\midrule
Conv2d(3,1), ReLU, BN & H $\times$ W $\times$ C &  (H/2) $\times$ (W/2) $\times$ (C$\times$2) & Conv-1\\
Conv2d(3,1), ReLU, BN & (H/2) $\times$ (W/2) $\times$ (C$\times$2) &  (H/4) $\times$ (W/4) $\times$ (C$\times$4) & Conv-2\\
Conv2d(3,1), ReLU, BN & (H/4) $\times$ (W/4) $\times$ (C$\times$4) &  (H/8) $\times$ (W/8) $\times$ (C$\times$8) & - \\
DeConv2d(3,1), ReLU, BN & (H/8) $\times$ (W/8) $\times$ (C$\times$8) &  (H/4) $\times$ (W/4) $\times$ (C$\times$4) & Deconv-1\\
Concatenate(Deconv-1, Conv-2) & - & (H/4) $\times$ (W/4) $\times$ (C$\times$8) & - \\
DeConv2d(3,1), ReLU, BN & (H/4) $\times$ (W/4) $\times$ (C$\times$8) &  (H/2) $\times$ (W/2)$\times$ (C$\times$2) & Deconv-2\\
Concatenate(Deconv-2, Conv-1) & - & (H/4) $\times$ (W/4) $\times$ (C$\times$4) & - \\
DeConv2d(3,1), ReLU, BN & (H/2) $\times$ (W/2) $\times$ (C$\times$4) & H $\times$ W $\times$ (C$\times$2) & Deconv-3\\
Concatenate(Deconv-3, Input image) & - & (H/4) $\times$ (W/4) $\times$ (C$\times$2+3) & - \\
Conv2d(3,1), ReLU, BN & H $\times$ W $\times$ (C$\times$2+3) &  H $\times$ W $\times$ C & - \\
Conv2d(3,1), tanh & H $\times$ W $\times$ C & H $\times$ W $\times$ ${\rm C_{out}}$ & - \\
\bottomrule
\end{tabular}
\caption{The network architecture of the residue predictor that decomposes an input image into color layers. Specifically, for decomposition into $7$ color layers, the residue predictor takes as inputs an image and $7$ RGBA palette layers of size $H \times W \times 4$ ($H \times W \times 3$ palette colors stacked on $H \times W \times 1$ processed alpha layers). Therefore, the total number of input channels is $C = 3 + 7 \times 4$. The output, composed of $7$ RGB layers, has ${\rm C_{out}} = 7 \times 3$ channels.}
\label{table:residue-predictor}
\end{center}
\end{table*}

\end{document}